%% file: main.tex
\documentclass{sig-alternate-2013}

\usepackage{epsfig}
\usepackage{graphicx}
\usepackage{amsmath}
\usepackage{amssymb}
\usepackage{multirow}
\usepackage{color}

\usepackage{fancyhdr}
\usepackage[breaklinks=true,bookmarks=false]{hyperref}

\newcommand{\todo}[1]{}
\renewcommand{\todo}[1]{{\color{red} TODO: {#1}}}

\permission{\copyright 2017 International World Wide Web Conference Committee \\ (IW3C2), published under Creative Commons CC BY 4.0 License.}
\conferenceinfo{WWW'17 Companion,}{April 3--7, 2017, Perth, Australia.}
\copyrightetc{ACM \the\acmcopyr}

\newcommand*{\refname}{Bibliography}

\begin{document}





\title{Visual Discovery at Pinterest}

\author{Andrew Zhai${^{^1\thanks{indicates equal contribution.}}}$, Dmitry Kislyuk${^{^1\footnotemark[1]}}$, Yushi Jing${^{^1\footnotemark[1]}}$, Michael Feng${^{^{1}}}$ \\ Eric Tzeng${^{^{12}}}$, Jeff Donahue${^{^{12}}}$,  Yue Li Du${^{^{1}}}$, Trevor Darrell${^{^{2}}}$ \\ \affaddr{${^{^1}}$ Visual Discovery, Pinterest} \qquad \affaddr{${^{^2}}$University of California, Berkeley} \\
\{andrew,dkislyuk,jing,m,etzeng,jdonahue,shirleydu\}@pinterest.com \\ trevor@eecs.berkeley.edu }


\maketitle


\input{0-abstract}






\input{1-intro}

\input{2-arch}

\input{3-applications}


\input{4-conclusion}

\bibliographystyle{abbrv}
\bibliography{main}
%





\end{document}

%% file: 0-abstract.tex
\begin{abstract}

Over the past three years Pinterest has experimented with several visual search and recommendation services, including Related Pins (2014), Similar Looks (2015), Flashlight (2016) and Lens (2017). This paper presents an overview of our visual discovery engine powering these services, and shares the rationales behind our technical and product decisions such as the use of object detection and interactive user interfaces. We conclude that this visual discovery engine significantly improves engagement in both search and recommendation tasks.

\end{abstract}

%% file: 1-intro.tex
\section{Introduction}

Visual search and recommendations~\cite{Datta:2008}, collectively referred to as \textit{visual discovery} in this paper, is a growing research area driven by the explosive growth of online photos and videos. Commercial visual search systems such as Google Goggles and Amazon Flow are designed to retrieve photos with the exact same object instance as the query image. On the other hand, recommendation systems such as those deployed by Google Similar Images~\cite{Wang:2014}, Shopping~\cite{bertelli2011kernelized} and Image Swirl~\cite{jing2012google} display a set of visually similar photos alongside the query image without the user making an explicit query.

\begin{figure}
	\centering \includegraphics[width=2.2 in]{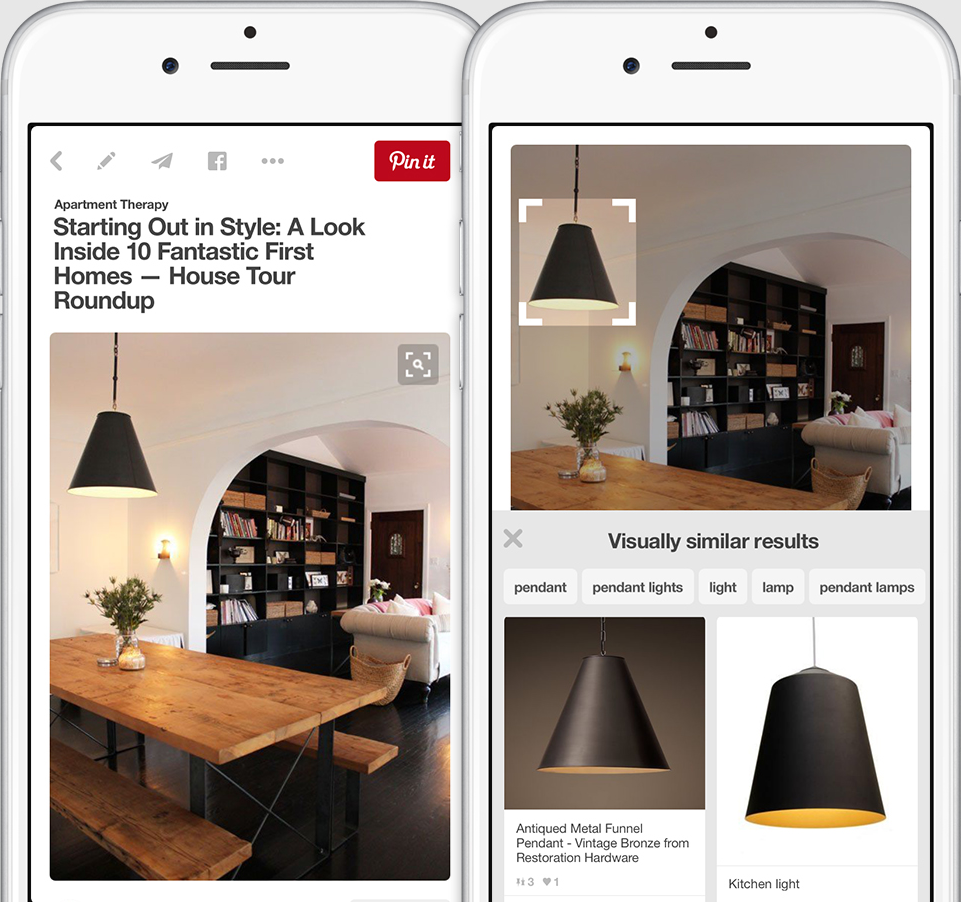}
	\caption{Pinterest Flashlight:  User can select any objects in the image (e.g. lamp, desk, shelf) as a visual search query.}
 	\label{fig:objectsearch1}
\end{figure}

At Pinterest we experimented with different types of visual discovery systems over a period of three and half years. We benefited from the confluence of two recent developments -- first, advances in computer vision, especially the use of convolutional networks and GPU acceleration, have led to significant improvements in tasks such as image classification and object detection. Second, a substantial number of users prefer using discovery systems to \textit{browse} (finding inspirational or related content) rather than to \textit{search} (finding answers). 
With hundreds of millions of users (who are often browsing for ideas in fashion, travel, interior design, recipes, etc.), Pinterest is a unique platform to experiment with various types of visual discovery experiences.

\begin{figure}
	\centering \includegraphics[width=3.2 in]{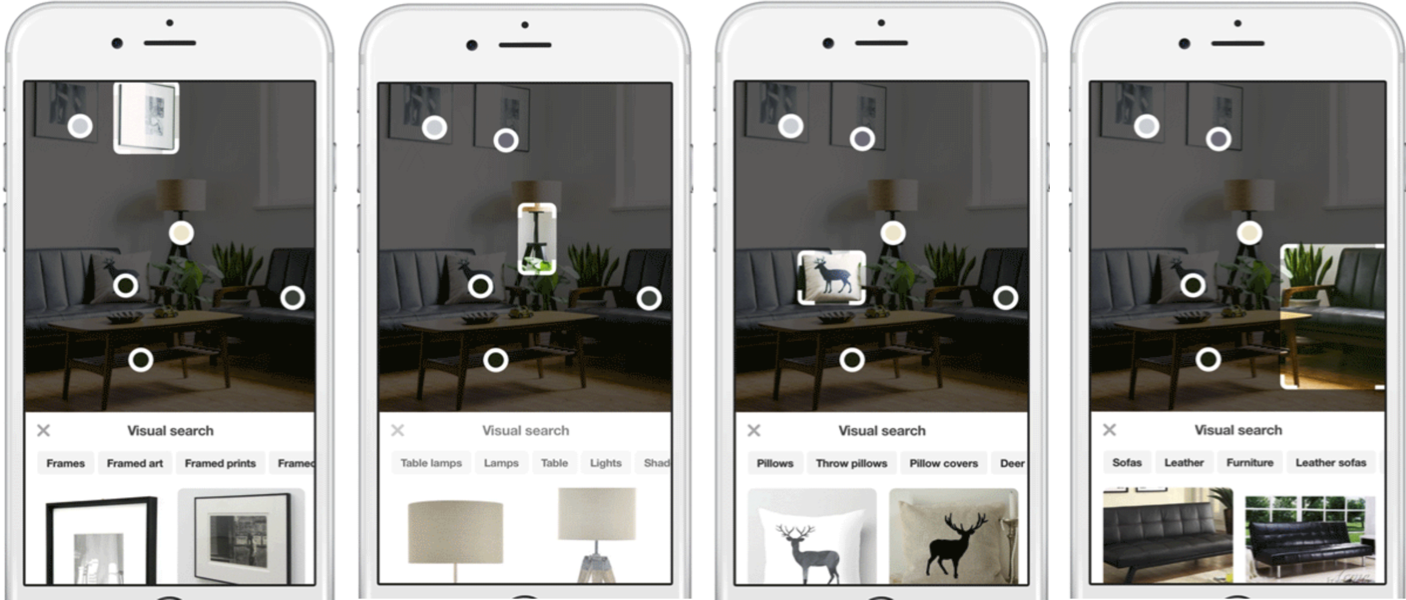}
	\caption{If objects are automatically detected, Flashlight displays a clickable ``dot'' for faster navigation.}
 	\label{fig:objectsearch4}
\end{figure}

\begin{figure}
	\centering \includegraphics[width=3.3 in]{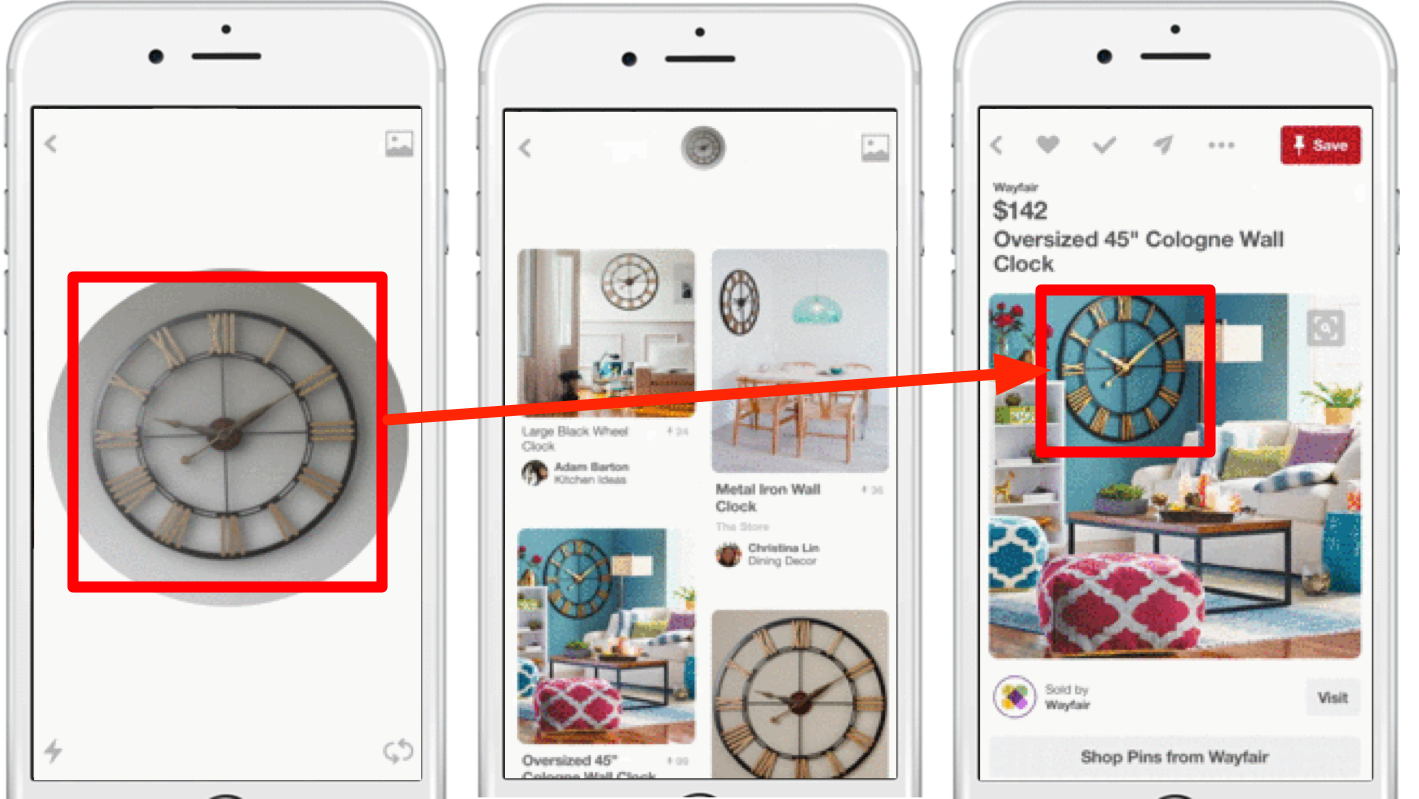}
	\caption{By indexing objects instead of whole images in our visual search system, Pinterest Lens findings objects \textit{within} the images.}
	\label{fig:reverse}
\end{figure}

Our previous work~\cite{jing15} presented preliminary results showing that convnet features and object detection can be used effectively to improve user engagement in visual search systems. We subsequently launched Pinterest Flashlight~\cite{flashlight}, an interactive visual discovery tool that allows user to select any object in the image (e.g. lamp, desk, shelf) as visual queries, as shown in Figure~\ref{fig:objectsearch1}.  If the system is certain about the location of an object, either through automatic object detection~\cite{rcnn} or collaborative filtering, Flashlight displays a clickable ``dot'' on the image for faster navigation, as shown in Figure~\ref{fig:objectsearch4}. Recently we launched Pinterest Lens~\cite{discovery}, a visual discovery experience accessed via a mobile phone camera, as shown in Figure~\ref{fig:reverse}. In this release we also applied object detection to the billions of images on Pinterest so that query objects can be matched against other visually similar objects \textit{within} the catalog images.
\newline\newline
This paper gives a general architectural overview of our visual discovery system and shares the lessons we learned from scaling and integrating visual features into products at Pinterest.
For example, we investigate the performance of popular classification networks for retrieval tasks and evaluate a binarization transformation~\cite{agrawal14analyzing} to improve the retrieval quality and efficiency of these classification features. This feature transformation enables us to drastically reduce the retrieval latency while improving the relevance of the results in large-scale settings. We also describe how we apply object detection to multiple visual discovery experiences including how to use detection as a feature in a general image recommendation system and for query normalization in a visual search system. 

The rest of the paper is organized as follows: Section~\ref{pinterest} and~\ref{related} give a brief introduction on the Pinterest image collection essential to our visual discovery experience and survey related works. Section~\ref{sec:features} and~\ref{sec:detection} present the visual features and object detectors used in our system. Section~\ref{sec:RelatedPins} describes how visual features can be used to enhance a generic image recommendation system. Section~\ref{sec:Flashlight} and~\ref{sec:lens} presents our experience launching Pinterest Flashlight and our latest application, Pinterest Lens. 


\section{Pinterest Images} 
\label{pinterest}
\begin{figure}
	\centering \includegraphics[width=3.3 in]{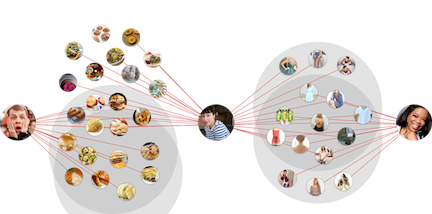}
	\caption{Pinterest users collect images into varieties of themed boards.}
	\label{fig:user2pin2board}
\end{figure}

Our work benefited significantly from having access to billions of catalog photos. On Pinterest, hundreds of millions of users organize images and videos around particular topics into boards as shown in Figure~\ref{fig:user2pin2board}, which result in a very large-scale and hand-curated collection with a rich set of metadata. Most Pin images are well annotated: when a person bookmarks an image on to a board, a Pin is created around an image and a brief text description supplied by the user. When the same Pin is subsequently saved to a new board by a different user, the original Pin gains additional metadata that the new user provides.  This data structure continues to expand every time an image is shared and saved. Such richly annotated datasets gave us the ability to generate training sets that can potentially scale to hundreds of thousands of object categories on Pinterest.

\section{Related Work}

\label{related}
\textbf{Visual discovery systems:} Amazon Flow, Google and Bing Similar Images are examples of widely used visual search and recommendation systems. Various works~\cite{visualrank}~\cite{distance08} have been proposed to improve the ranking of the image search results using visual features.  In recent years, research has also focused on domain-specific image retrieval systems such as fashion recommendation~\cite{yamaguchi}~\cite{liu2016deepfashion}~\cite{Simo-Serra_2016_CVPR}, product recommendation~\cite{streetshop}~\cite{WhereToBuyItICCV15} and discovery of food images~\cite{AizawaO15}. Compared with existing commercial visual discovery engines, our system focuses more on \textit{interactive} retrieval of objects within the images.
\newline
\newline
\textbf{Representation learning:} Over the past few years, convnet architectures such as AlexNet~\cite{alexnet12}, GoogLeNet~\cite{googlenetv4}, VGG and ResNet have continuously pushed the state of the art on large-scale image classification challenges. Though trained for classification in one particular domain, the visual features extracted by these models perform well when transferred to other classification tasks~\cite{cnntransfer}, as well as related localization tasks like object detection and semantic segmentation~\cite{jonevan}. This paper presents empirical evaluation of widely used classification-based image features in retrieval settings.~\footnote{ Recent work has also demonstrated the effectiveness of learning embeddings or distance functions directly from ranking labels such as relative comparisons~\cite{facenet}~\cite{queryspecific} and variations of pairwise comparisons~\cite{Bell:2015}~\cite{songCVPR16}, or using Bilinear features~\cite{compactbilinear} for fine-grained recognition. We will report our empirical findings in our future work.}
\newline
\newline
\textbf{Detection:} For web-scale services, the part-based model~\cite{partsbased10} approach was a well studied for detection, but recently deep learning methods have become more prevalent, with applications such as face detection~\cite{deepface}, street number detection~\cite{google_street}, and text detection~\cite{readingtext}. Recent research focuses on application of detection architectures such as Faster R-CNN~\cite{ren2015faster}, YOLO~\cite{Redmon_2016_CVPR} and the Single Shot Detector~\cite{liu2016ssd}. In this work we present, to the best of our knowledge, the first end-to-end object detection system for large-scale visual discovery systems.

%% file: 2-arch.tex
\input{2.1-features}

\input{2.2-detection}

%% file: 2.1-features.tex
\section{Feature Representation}
\label{sec:features}

We adopted and evaluated several popular classification models such as AlexNet\cite{alexnet12}, GoogLeNet~\cite{googlenet2014}, VGG16~\cite{vggnet2014}, and variants ResNet101 and ResNet152\cite{kaiming16}.  In addition to the raw features, we also explored binarized~\cite{agrawal14analyzing} representations of these features which are of interest to us due to their much smaller memory footprint. We also compare the Euclidean (L2) and Manhattan (L1) distance metrics for the raw features. For these features, training and inference are done through the 
open-source Caffe~\cite{jia2014caffe} framework on multi-GPU machines. We are separately investigating on-device visual discovery experience to augment the server-based approach using the mobile implementation of TensorFlow~\cite{tensorflow}. 

Beyond the base models trained for ImageNet classification~\cite{imagenet_cvpr09}, we also fine-tune on our own Pinterest training dataset generated in a similar fashion as the evaluation dataset while ensuring no overlap. This is because Pinterest images, many of which are high-quality stock photography and professional product images, have different statistics than ImageNet images, many of which are personal photos focused on individual generic objects.
Models are fine-tuned by replacing the softmax classification layer of a pre-trained base model with a new classification layer, trained to classify Pinterest images, initialized using the same intermediate and lower level weights.


\begin{table}
\centering
\caption{Precision@K on Pinterest evaluation dataset.}
\label{table:visualsearchrel}
\resizebox{\columnwidth}{!}{%

\begin{tabular}{ | l | c | c | c | c | c | c | } \hline
 & & & & & & \\
\textbf{Model}  & \textbf{Layer} & \textbf{Type} & \textbf{Dist.} & \textbf{P@1} &  \textbf{P@5}  &  \textbf{P@10}   \\
\hline
AlexNet     & fc6              & raw & L2 & 0.093 & 0.045 & 0.027 \\
AlexNet    & fc6              & raw & L1 & 0.099 & 0.047 & 0.027 \\
AlexNet     & fc7              & raw & L2 & 0.104 & 0.052 & 0.031 \\
AlexNet      & fc7              & raw & L1 & 0.106 & 0.052 & 0.031 \\
AlexNet     & fc8              & raw & L2 & 0.088 & 0.047 & 0.028 \\
AlexNet      & fc8              & raw & L1 & 0.090 & 0.046 & 0.028 \\
GoogleNet    & loss3/classifier & raw & L2 & 0.095 & 0.050 & 0.032 \\
GoogleNet      & loss3/classifier & raw & L1 & 0.098 & 0.050 & 0.032 \\
VGG16          & fc6              & raw & L2 & 0.108 & 0.051 & 0.030 \\
VGG16           & fc6              & raw & L1 & 0.118 & 0.057 & 0.035 \\
VGG16          & fc7              & raw & L2 & 0.116 & 0.058 & 0.036 \\
VGG16           & fc7              & raw & L1 & 0.113 & 0.060 & 0.038 \\
VGG16           & fc8              & raw & L2 & 0.104 & 0.054 & 0.034 \\
VGG16           & fc8              & raw & L1 & 0.106 & 0.054 & 0.034 \\
ResNet101 & pool5  & raw & L2 & 0.160 & 0.080 & 0.050 \\
ResNet101 & pool5  & raw & L1 & 0.149 & 0.073 & 0.045 \\
ResNet101 & fc1000 & raw & L2 & 0.133 & 0.068 & 0.042 \\
ResNet101 & fc1000 & raw & L1 & 0.139 & 0.067 & 0.041 \\
\textbf{ResNet152} & pool5  & raw & L2 & \textbf{0.170} & \textbf{0.083} & \textbf{0.050} \\
ResNet152 & pool5  & raw & L1 & 0.152 & 0.077 & 0.047 \\
ResNet152 & fc1000 & raw & L2 & 0.149 & 0.073 & 0.045 \\
ResNet152 & fc1000 & raw & L1 & 0.148 & 0.073 & 0.044 \\
\hline
AlexNet     & fc6              & binary & H. & 0.129 & 0.065 & 0.039 \\
AlexNet     & fc7              & binary & H. & 0.110 & 0.054 & 0.033 \\
AlexNet      & fc8              & binary & H. & 0.089 & 0.046 & 0.027 \\
\textbf{VGG16} & fc6 & binary & H. & \textbf{0.158} & \textbf{0.081} & \textbf{0.049} \\
VGG16           & fc7              & binary & H. & 0.133 & 0.068 & 0.044 \\
VGG16            & fc8              & binary & H. & 0.110 & 0.055 & 0.035 \\
ResNet101 & fc1000 & binary & H. & 0.125 & 0.062 & 0.039 \\
ResNet101 & pool5  & binary & H. & 0.055 & 0.025 & 0.014 \\
ResNet152 & fc1000 & binary & H. & 0.133 & 0.065 & 0.041 \\
ResNet152 & pool5  & binary & H. & 0.057 & 0.026 & 0.015 \\
\hline
\textbf{VGG16 (Pin.)}   & fc6              & binary & H. & \textbf{0.169} & \textbf{0.089} & \textbf{0.056} \\

\hline
\end{tabular}
}
\end{table}

\begin{figure}
\vspace*{-0.7cm}
\centering
\includegraphics[width=3.2 in]{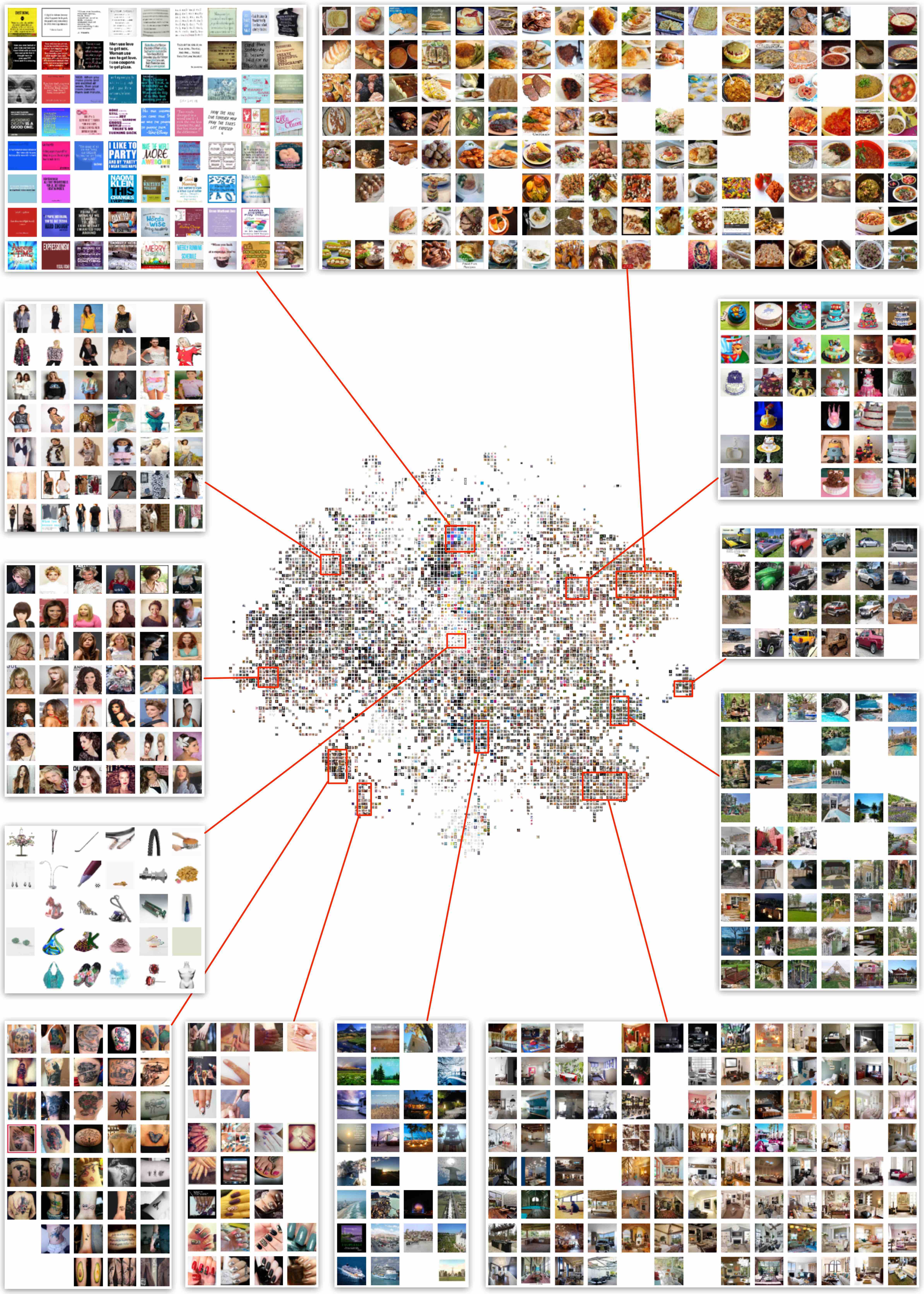}
\caption{Visualization of binarized embeddings of Pinterest images extracted from fine-tuned VGG16 FC6 layer.}
\label{fig:embeddings_tsne}
\end{figure}

The training dataset consists of millions of images distributed over roughly 20,000 classes while the evaluation dataset, visualized in Figure~\ref{fig:embeddings_tsne}, consists of 100,000 images distributed over the same classes. Both datasets were collected from a corpus of Pinterest images that are labeled with annotations. We limit the set of annotations by taking the top 100,000 text search queries on Pinterest and randomly sampling 20,000 of them. After filtering, we retain a corpus of images containing only annotations matching the sampled queries and randomly sample from this corpus to create balanced training and evaluation datasets. The query annotation used to generate the image becomes its class label.

To evaluate our visual models, from our evaluation dataset we use 2,000 images as query images while the rest are indexed by our visual search system using the image representation being evaluated. We then retrieve a list of results for each query image, sorted by distance. A visual search result is assumed to be relevant to a query image if the two images share the same class label, an approach that is commonly used for offline evaluation of visual search systems~\cite{Muller:2001} in addition to human evaluation. From the list of results with the associated class labels, we compute Precision @ K metrics to evaluate the performance of the visual models.

The results of these experiments for P@1, P@5, and P@10 performance are shown in Table~\ref{table:visualsearchrel}. We see that when using raw features, intermediate features (pool5, fc6, fc7) result in better retrieval performance than more semantic features (fc8, fc1000).
Among the raw feature results, ResNet152 pool5 features using L2 distance perform the best. For scalability, however, we are most interested in performance using binary features as our system must scale to billions of images. When binarized, VGG16 fc6 features perform best among the features we tested. These binarized fc6 features are 16x smaller than the raw pool5 features, with a trade-off in performance.  Fine-tuning the VGG16 weights on Pinterest data, however, makes up for this performance difference.

\begin{figure}
\centering
\includegraphics[width=3.2 in]{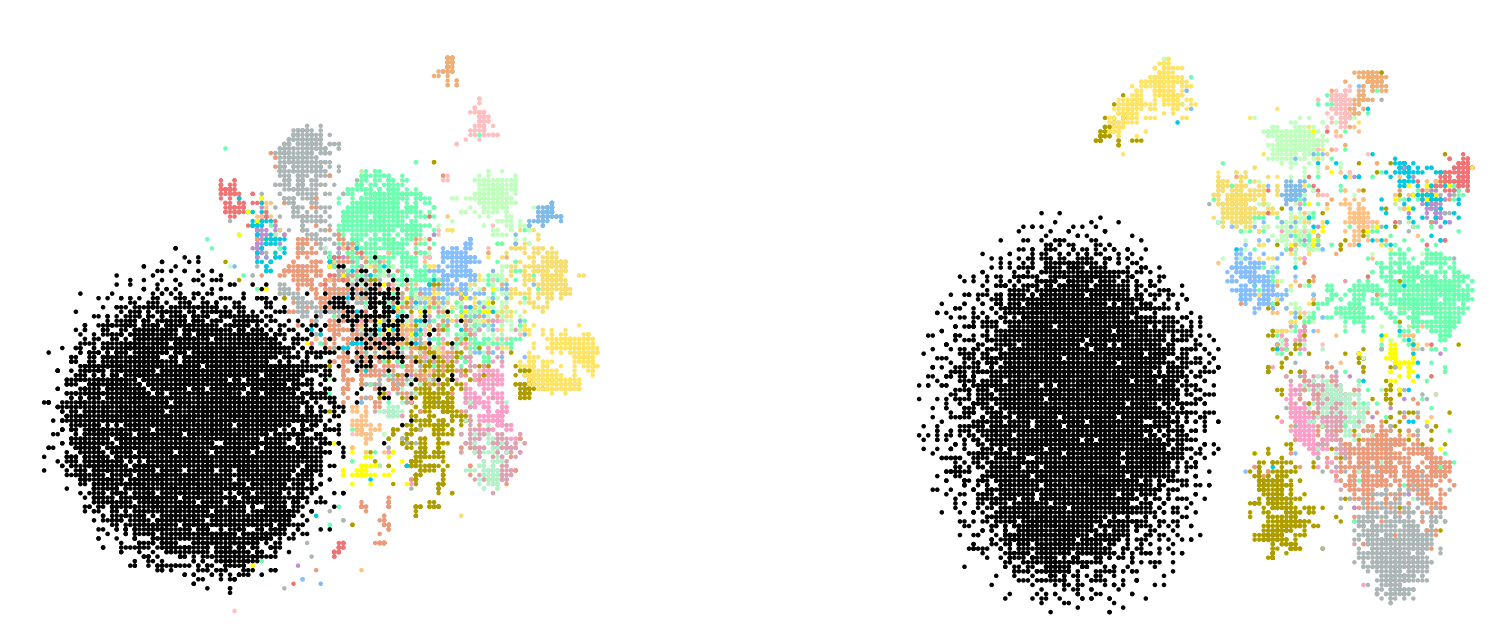}
\caption{VGG16 fc6 layer raw features (left) vs binary features (right). PASCAL VOC 2011 classification images are shown as colored dots. White noise images are shown as black dots. One can see that binarization separates out the noise from the Pascal images while retaining the label clusters.}
\label{fig:binarization}
\end{figure}


\begin{table}[]
\centering
\caption{Applying binary transformation to the VGG16 fc6 feature improves precision@k on 
PASCAL VOC 2011 images}
\label{table:binary_vs_raw}
\begin{tabular}{@{}llllll@{}}
\hline
  \textbf{Type}   & \textbf{Dist} & \textbf{P@1}   & \textbf{P@5}   & \textbf{P@10}  \\ \hline
raw    & L2 & 0.588 & 0.544 & 0.506 \\
 raw    & L1 & 0.658 & 0.599 & 0.566 \\
\textbf{binary} & \textbf{H.}   & \textbf{0.752} & \textbf{0.718} & \textbf{0.703} \\
\end{tabular}
\end{table}

We conjecture that the drop in performance of the ResNet pool5 feature when binarized is due to how the pool5 features are average pooled from ReLU activations. ReLU activations are either zero or positive and averaging these activation will bias towards positive pool5 features that are each binarized to one. We see that for ResNet101 pool5 features on our evaluation dataset, on average 83.86\% of the features are positive. Additionally, we speculate that intermediate features of AlexNet and VGG16 are better suited for the current binarization scheme. By training the intermediate features with a ReLU layer, the network will learn features that ignore the magnitude of negative activations. We see that this is also exactly what the binarization post-processing does for negative features. 

One interesting observation is that binarization improves the retrieval performance of AlexNet and VGG16. We reproduce this behavior in Table ~\ref{table:binary_vs_raw} on the Pascal VOC 2011~\cite{pascal-voc-2011} classification dataset. In this experiment, we built our index using fc6 features extracted from VGG16 from the train and validation images of the Pascal classification task. Each image is labeled with a single class with images that have multiple conflicting classes removed. We randomly pick 25 images per class to be used as query images and measure precision@K with $K = 1, 5, 10$. With the Pascal dataset, we can also qualitatively see that one advantage of binarization is to create features that are more robust to noise. We see this effect in Figure ~\ref{fig:binarization} where binarization is able to cleanly separate noisy images from real images while the raw features cluster both together.

%% file: 2.2-detection.tex
\section{Object Detection}
\label{sec:detection}

One feature that is particularly relevant to Pinterest is the presence of certain object classes (such as shoes, chairs, tables, bags, etc.). Extracting objects from images not only allows us to build novel discovery experiences (e.g. Section~\ref{sec:lens}), but also improves user engagement metrics as detailed in Sections~\ref{sec:RelatedPins} and ~\ref{sec:Flashlight}. This section covers our iterations of object detection at Pinterest, starting from deformable parts models described in our prior work~\cite{jing15} to Faster R-CNN and Single Shot Detection (SSD) described below.
\newline

\textbf{Faster R-CNN}: One approach of interest for object detection is Faster R-CNN, given its state-of-the-art detection performance, reasonable latency for real-time applications~\cite{ren2015faster}, and favorable scalability with a high number of categories (since the vast majority of parameters are shared). We experimented with Faster R-CNN models using both VGG16~\cite{vggnet2014} and ResNet101~\cite{kaiming16} as base architectures and the ResNet101 variant is currently one of the detection models, along with SSD, used in production at Pinterest.


When training Faster R-CNN models, we make use of a few differences from the original presentation of the method.
First, we train our models using direct end-to-end optimization, rather than the alternating optimization presented in the original paper.
Additionally, for the ResNet101 architecture of Faster R-CNN, we found that significant computation savings could be achieved by decreasing the number of proposals considered from 300 to 100 (or dynamically thresholded based on the objectness score of the ROI) during inference without impacting precision or recall. The shared base convolution features are pre-trained on a fine-tuned Pinterest classification task, and then trained for 300k iterations using an internal Pinterest detection dataset.
\newline
\newline
\textbf{Single Shot Detection:} Although Faster R-CNN is considerably faster than previous methods while achieving state-of-the-art precision, we found that the latency and GPU memory requirements were still limiting. For example, the productionized implementation of Faster R-CNN used in Flashlight, as described in Section~\ref{sec:Flashlight}, relies on aggressive caching, coupled with dark reads during model swaps for cache warming. We also index objects into our visual search system to power core functionalities of Lens, as described in Section~\ref{sec:lens}, which requires running our detector over billions of images. Any speedups we can get in such use cases can lead to drastic cost savings. As a result, we have also experimented with a variant of the Single Shot Multibox Detector (SSD)~\cite{liu2016ssd}, a method known to give strong detection results in a fraction of the time of Faster R-CNN-based models.  A more comprehensive evaluation on the speed/accuracy trade-off of object detection is described in ~\cite{detection_tradeoff}.

Because speed is one of our primary concerns, we perform detection with SSD at a relatively small resolution of $290 \times 290$ pixels.
Our VGG-based architecture resembles that of the original authors very closely, so we refer the reader to the original work for more details.
Here, we outline a few key differences between our model and that of the original authors.

\begin{figure}
\centering
\includegraphics[width=\columnwidth]{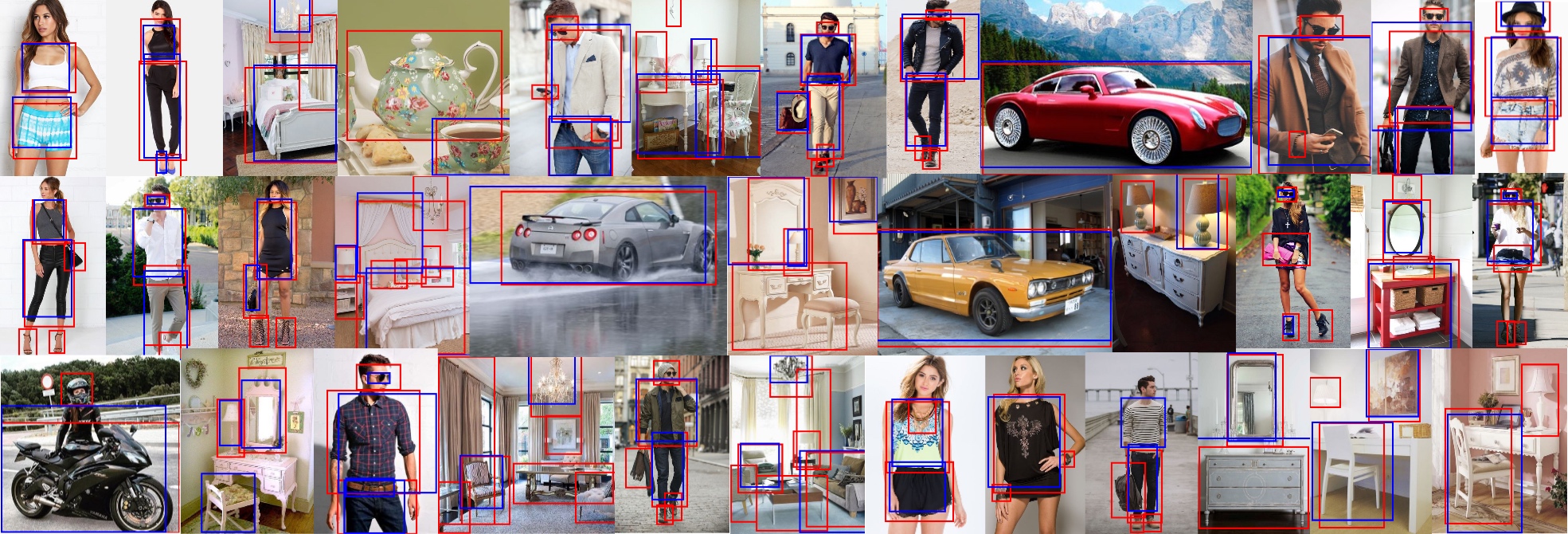}
\caption{Sample object detection results from the Faster R-CNN model (labeled as red) and SSD model (labeled as blue)}
\end{figure}

\input{tables/detection}

First, we train our models with a 0.7 IoU threshold instead of a 0.5 threshold, to ensure that the resulting detections tightly bound the objects in question.
The original SSD architecture uses additional convolutional layers with strides of 2 to detect larger objects---however, we found that this led to poor localization performance under our desired 0.7 IoU threshold.
Instead, our model to uses a stride of 1 for all additional convolutions, which greatly improves the quality of bounding box localization.
Due to the size of our detection dataset (74,653 images), we also found that randomly sampling positive and negative anchors in a 1:3 ratio, as in the original paper, led to poor convergence.
Instead, we make use of the recent Online Hard Example Mining (OHEM) method to sample anchors by always selecting the anchors that the models incurs the largest loss for~\cite{shrivastava2016ohem}.
However, despite our success with OHEM, we note that OHEM actually led to overfitting on a smaller detection dataset (19,994 images), indicating that OHEM is perhaps most useful on large and difficult datasets.

With these changes, we were able to train an SSD-based model that can perform object detection on $290 \times 290$ images in just 93 ms on an NVIDIA GRID K520 (an older generation GPU supported by Amazon Web Services), while achieving an F1 score on our internal evaluation dataset comparable to the previously described ResNet Faster R-CNN model.


Table ~\ref{table:detection} provides precision, recall, and latency figures on a Pinterest detection evaluation dataset, using our Faster R-CNN and SSD implementations. This dataset contains 79 categories, broadly categorized by \textit{Fashion}, \textit{Home Decor}, and \textit{Vehicles}. This evaluation was performed at a 0.7 IoU threshold, and although the precision results look favorable for SSD, we did observe, qualitatively, that the localization quality was worse (one reason being that SSD uses a smaller, warped input size). Nevertheless, SSD, along with the newer iteration of the YOLO model, owning to their simpler and more streamlined architectures (and superior latency), warrant close investigation as the favored object detection models for production applications.

%% file: tables/detection.tex
\begin{table}
	\centering
	\caption{Object detection performance.}
	\label{table:detection}
	\begin{tabular}{ c | cc | cc }

		 & \multicolumn{2}{c |}{\textbf{Faster R-CNN}} & \multicolumn{2}{c}{\textbf{SSD}} \\
			& precision & recall & precision & recall \\
		\hline
	  \textbf{Fashion} &	0.449     & 0.474     &  0.473     & 0.387                  \\
	  \textbf{Home decor} & 0.413     & 0.466    & 0.515     & 0.360              \\ 
	  \textbf{Vehicles} & 0.676     & 0.625     & 0.775     & 0.775           \\ 
	  \hline
	  \textbf{Overall} & 0.426     & 0.470  & 0.502     & 0.371  \\
	  \hline
	  \textbf{Latency}  & \multicolumn{2}{c |}{272 ms } & \multicolumn{2}{c}{59 ms} \\
	\end{tabular}
	                                                       
\end{table}

%% file: 3-applications.tex
\input{3.1-related-pins}
\input{3.3-object-search}
\input{3.4-future}

%% file: 3.1-related-pins.tex
\section{Pinterest Related Pins}
\label{sec:RelatedPins}

\begin{figure}
	\centering \includegraphics[width=3.0 in]{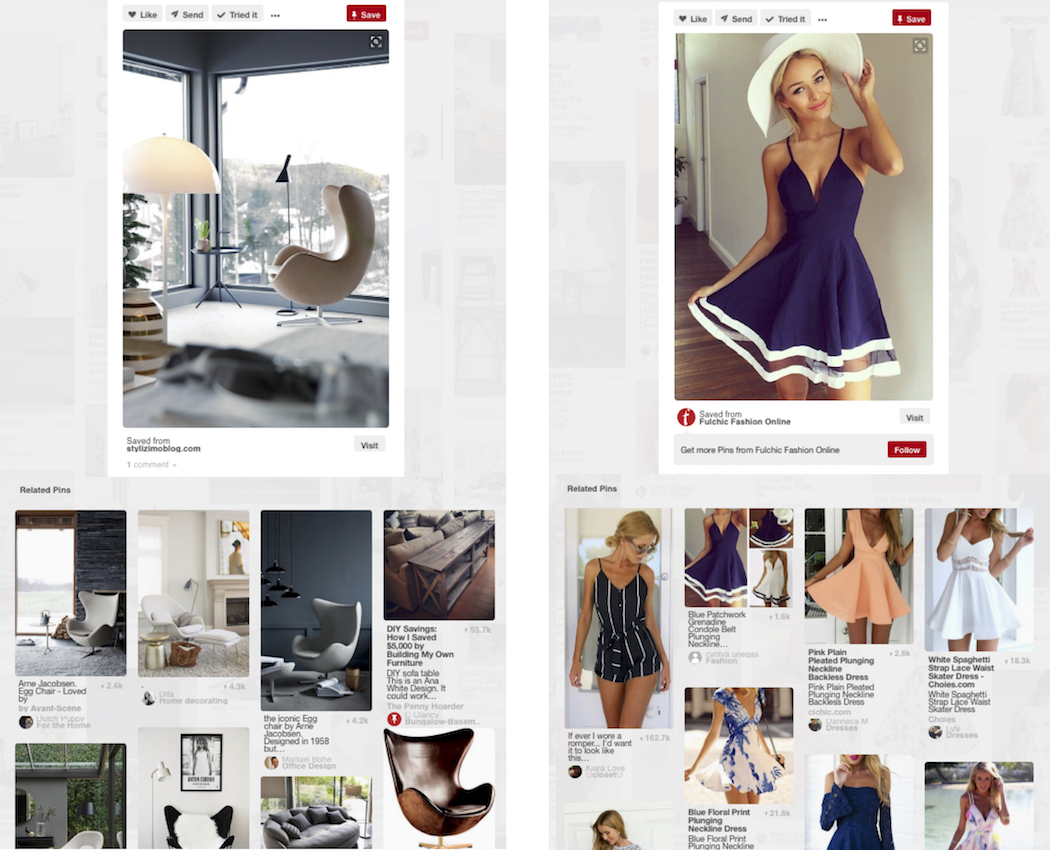}
	\caption{Related Pins is an item-to-item recommendation system.  The results are displayed below the currently viewing image.}
	\label{fig:related}
\end{figure}

Related Pins is a pin recommendation system that leverages the vast amount of human-curated content on Pinterest to provide personalized recommendations of pins based on a
given query pin. It is most heavily used on the pin closeup
view shown in Figure~\ref{fig:related}, which is known as the Related Pins
feed. Additionally, Related Pins recommendations have been
incorporated into several other parts of Pinterest, including
the home feed, pin pages for unauthenticated visitors, emails,
and certain curated pin collections (such as the Explore tab).

User engagement on Pinterest is defined by the following
actions. A user closeups on a pin by clicking to see more
details about the pin. The user can then click to visit the
associated Web link; if they remain on-site for an extended
period of time, it is considered a long click. Finally, the user
can save pins onto their own boards. We are interested in
driving ``Related Pins Save Propensity" which is defined as
the number of users who have saved a Related Pins recommended
pin divided by the number of users who have seen a
Related Pins recommended pin. Liu et al.~\cite{relatedpins} presented a more detailed architecture overview and evolution of the related pins feature. This section focuses on how convnet features and object detection can improve the engagement of Related Pins. 


\subsubsection*{Covnet features for recommendations}

\label{sec:app-related-pins}

Related Pins are powered through collaborative filtering (via image-board co-occurrences) for candidate generation and the commonly used Rank-SVM to learn a function where various input features are used to generate a score for re-ranking. To conduct the experiment, we set up a series of A/B experiments, where we selected five million popular Pins on Pinterest as queries, and re-ranked their recommendations using different sets of features. The control group re-ranked Related Pins using a linear model with the existing set of features. The treatment group re-ranked using fine-tuned VGG \textit{fc6} and \textit{fc8} visual similarity features along with indicator variables (in addition to the features used in control).


Across the 5M query Pins, the treatment saw a 3.2\% increase in engagement (click and repin) \footnote{This metric was measured across a 14 day period in Sep. 2015.}. After expanding the treatment to 100M query Pins, we observed a net gain of 4.0\% in engagement with Related Pins, and subsequently launched this model into production. Similar experiments with a fine-tuned AlexNet model yielded worse results (only 0.8\% engagement gain) as expected from our offline evaluation of our visual feature representation.

When broken down by category, we noted that the engagement gain was stronger in predominantly visual categories, such as art (8.8\%), tattoos (8.0\%), illustrations (7.9\%), and design (7.7\%), and lower in categories which primarily rely on text, such as quotes (2.0\%) and fitness planning (0.2\%). Given the difference in performance among categories, we performed a follow-up experiment where we introduced a cross feature between the category vector of the query and the scalar \textit{fc6} visual similarity feature (between the query and candidate) to capture the dependency between category and usefulness of the visual features in our linear model. This introduces 32 new features to the model, one for each of our site-wide categories (these features are sparse, since the Pinterest category vector thresholds most values to zero). The result from this was a further 1.2\% engagement increase in addition to the gains from the initial visual re-ranking model.

\subsubsection*{Object detection for recommendation}

To validate the effectiveness of our detection system in production, one of our first experiments was to further improve the use of visual features in Related Pins (described in the previous section). Our primary observation was that users are sometimes interested in the \textit{objects} in the Pin's image, instead of the full image, and we therefore speculated that object detection could help compute \textit{targeted} visual similarity features. For the below experiments, we focus on fashion queries as only our fine-tuned fashion Faster R-CNN model was available at the time of this experiment.

\begin{table}[h]
\centering
\caption{Engagement results when adding cross features and object detection to visual similarity feature, measured over a 7 day period in Oct. 2015.}
\resizebox{\columnwidth}{!}{%
\begin{tabular}{  l  |  l |  c }

   \textbf{Features}  &  \textbf{Queries}  & \textbf{Engagement}    \\
\hline
FT-VGG (control)	 &  - & - \\
\hline
FT-VGG + category cross features	 & 5M & +1.2\% \\
\hline
FT-VGG + object detection \textit{variant A}	 & 315k fashion  & +0.5\%  \\
FT-VGG + object detection \textit{variant B}	 & 315k fashion & +0.9\%  \\
FT-VGG + object detection \textit{variant C}	 & 315k fashion & +4.9\%  \\

\end{tabular}
}
\label{tbl:visualsearchrel}
\end{table}

After applying non-maximum suppression (NMS) to the proposals generated by our Faster R-CNN detector, we considered query Pins where the largest proposal either occupies at least 25\% of the Pin's image or if the confidence of the proposal passes a threshold of 0.9. We categorize these images as containing a \textit{dominant visual object}, and using the best-performing fine-tuned VGG re-ranking variant from the previous section as our control, we experimented with the following treatments: 
\begin{itemize}
\item \textit{Variant A}: if a dominant visual object is detected in the query Pin, we compute visual features (VGG fc6) on just that object.
\item \textit{Variant B}: same as \textit{variant A}, but we also hand-tune the ranking model by increasing the weight given to visual similarity by a factor of 5. The intuition behind this variant is that when a dominant visual object is present, visual similarity becomes more important for recommendation quality.
\item \textit{Variant C}: Both control and treatment have the same visual features. If the query contains a dominant visual object however, we increase the weight of the visual similarity features by a factor of 5, as in \textit{variant B}. In this variant, we assume that the presence of detected visual objects such as bags or shoes indicates that visual similarity is more important for this query. 
\end{itemize}
Results for these variants are listed in Table \ref{tbl:visualsearchrel}. Variants A and B of the object detection experiments show that directly computing visual features from object bounding boxes slightly increases engagement. One suspicion for the lack of more significant performance increases may be due to our detector returning tight bounding boxes that do not provide enough context for our image representation models as we are matching a query object to whole image candidates. Variant C shows a more significant 4.9\% engagement gain over the VGG similarity feature control, demonstrating that the presence of visual objects alone indicates that visual similarity should be weighed more heavily.

Our object detection system underwent significant improvements following the promising results of this initial experiment, and was applied to substantially improve engagement on Flashlight, as described in the next section.

%% file: 3.3-object-search.tex
\section{Pinterest Flashlight}
\label{sec:Flashlight}

\begin{figure*}
	\centering
	\includegraphics[width=5 in]{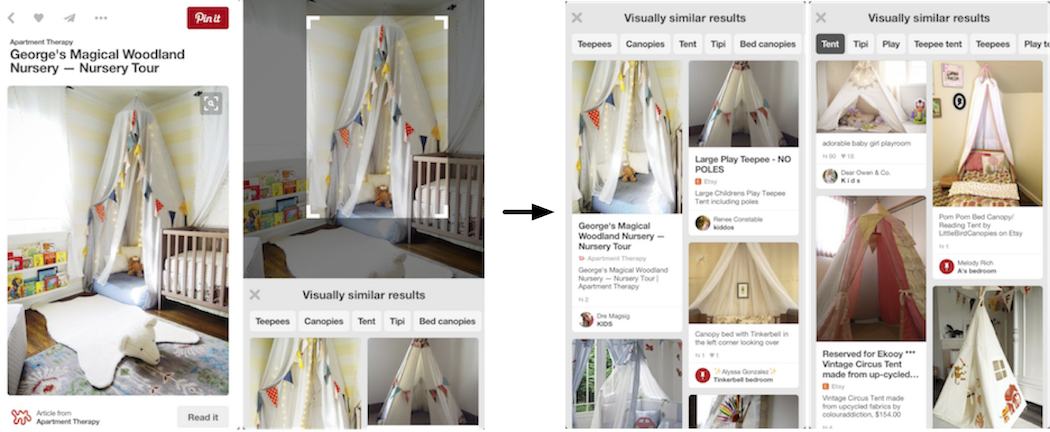}
	\caption{Pinterest Flashlight supports interactive retrieval of objects in the images through the use of ``cropper'' tool (left) and query-refinement suggestions (top right). }
	\label{fig:objectsearchflow_crop}
\end{figure*}

\begin{figure}
	\centering \includegraphics[width=3.3 in]{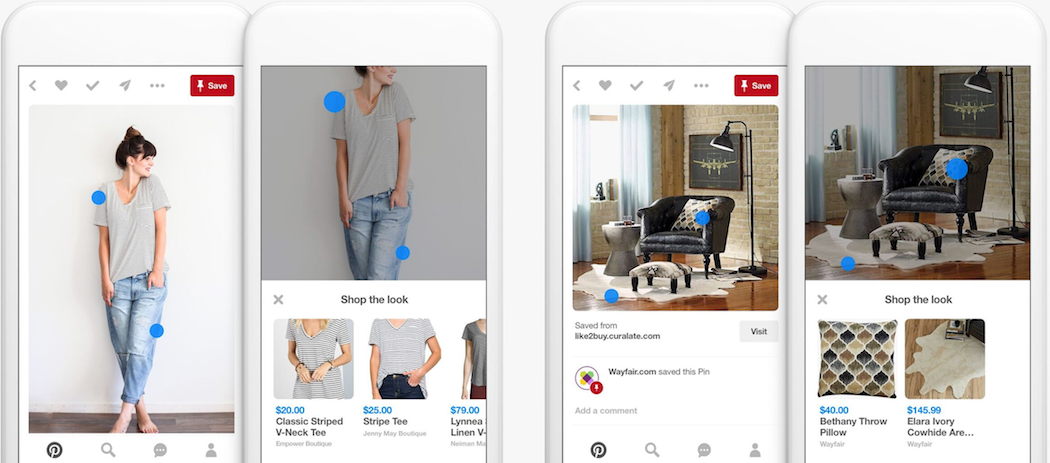}
	\caption{Shop-the-Look uses a variant of Flashlight returning only buyable content along with manual tagging to achieve a high quality visual shopping experience.}
 	\label{fig:similarlooks}
\end{figure}

Flashlight, as shown in Figure~\ref{fig:objectsearchflow_crop} is a visual search tool that lets users search for object within any image on Pinterest. It supports various applications including Shop-the-Look~\footnote{Shop-the-Look combines visual search results with expert curation to improve the accuracy of the results.}, which launched in 2017 and shown in Figure~\ref{fig:similarlooks}.

The input to Flashlight is an object proposal, which is generated either (1) using the detection features described in Section~\ref{sec:detection} or (2) directly from our users via the flexible cropping mechanism as shown in Figure~\ref{fig:objectsearchflow_crop}. Visual search on the object is then powered by the retrieval features described in Section~\ref{sec:features} for candidate generation with light weight reranking using the full image metadata as context. Flashlight returns both image results and clickable tf-idf weighted annotations to allow users to narrow down on results. These annotations are generated by aggregating the annotations of the image results.

We implemented Flashlight as an iteration of Similar Looks, a fashion specific visual search tool described in our prior work~\cite{jing15}. We learned from Similar Looks that being able to find similar results of objects within an image can increase engagement. However, after running the experiment, we saw challenges in launching the Similar Looks product into wider deployment. First, due to the offline nature of our older parts-based detection pipeline, the coverage of objects was too low. Only 2.4\% of daily active users saw a detected object and user research revealed that inconsistency with how some images had objects while other images did not confused users, who expected interactive dots on every image, including new content. 

The cropping mechanism was our solution to this problem, giving users the flexibility to manually select any object in any image and get real-time visually similar results for the selected object. By not restricting what a user can crop, any object can be searched. A few months after the initial launch, we introduced object detection to Flashlight, generating clickable dots to simplify the user interface. At the time of the publication, more than 1 billion instances of objects were detected and shown to the user.

\subsubsection*{Convnet Features for Retrieval}
The initial Flashlight launch did not include automatic object detection. We introduced a semi-transparent search icon on the top right of a Pin image that a users can press to use the cropping mechanism. As the user manipulates the crop bounding box, real time visual search results and annotations are generated. 

Initially after the launch we saw that 2\% of users who viewed a pin closeup will press the search button, which corresponds to roughly 4 million visual search request per day. Because search is done on crops of an image, when retrieving visually similar results, the re-ranking function normally applied after the candidate retrieval step is very minimal \footnote{Some components of the re-ranking function such as near-dupe image scoring still apply} as we have no metadata for the crop. Therefore, the visual search results returned are mostly determined by the initial nearest neighbor retrieval on our deep learning embeddings. With the context that it is mostly deep learning that is powering this new feature, we are able to achieve 2/3 the engagement rate of text search, an existing product whose quality has been in development for many years.


%

\subsubsection*{Object Detection for Retrieval}

\begin{table}
\begin{center}
\caption{Object Detection in Flashlight (Faster R-CNN ResNet101), detection threshold 0.7}
\label{tbl:spotlight_eng}
{
\begin{tabular}{  l  | l | l | l | l | l }
 Visual     & Annot. & Categ. & Engagement     \\
Thresh.  & Thresh. & Conformity &  Gain \\
\hline
0.0 & 0 & 0.0 & -3.8\% \\
1.0 & 0 & 0.0 & -5.3\% \\
 0.0 & 1000 & 0.0 & -5.2\% \\
1.0 & 1000 & 0.0 & +2.1\% \\
1.0 & 1000 & 0.8 & \textbf{+4.9}\% \\
\end{tabular}
}
\end{center}
\end{table}

One crucial improvement to Flashlight that we implemented is real-time object detection using Faster R-CNN, which offers several advantages to the product. First, we're able to simplify the user interface of Flashlight by placing predetermined \textit{object dots} for categories we detect so users could click on the dot instead of manually moving a cropping box. Second, a useful goal for any retrieval system is to construct an aggregated user engagement signal, one use case being to boost up known highly engaged results for a query. This aggregation was infeasible in our initial version of Flashlight as most queries are unique due to the use of a manual cropper.

\begin{figure}
	\includegraphics[width=3.4 in]{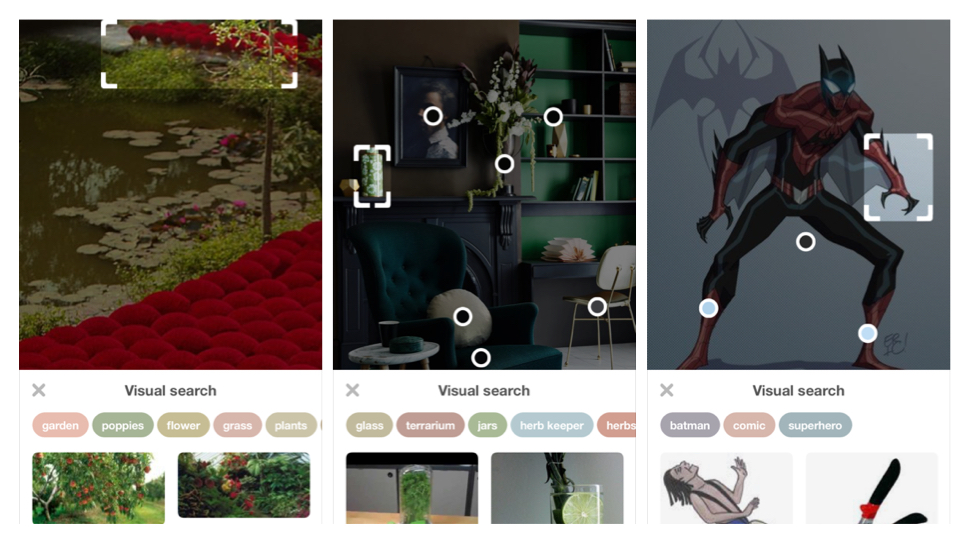}
	\caption{Examples of object detection false positives (shown as selected object), which are successfully suppressed using the category conformity suppression method.
     }
	\label{fig:object_fp}
\end{figure}

Though intuitively, object detection seemed to be a simple engagement improvement to Flashlight as users would have a much easier interface to use, we learned otherwise through our A/B experiments. Upon initial launch of the object detection experiment, where the control displayed a default bounding box, and treatment displayed a clickable dot over each detected object (as in Figure ~\ref{fig:objectsearch4}) we found that engagement metrics decreased (specifically, we were interested in ``Flashlight Save Propensity" similar to the Related Pins metric used previously). After investigation into the reason for the poor performance, We saw two significant issues, as shown in Figure \ref{fig:object_fp}: (a) bad object detections (manifested as either low localization quality or object ``hallucination", left example), and (b) irrelevant retrieval results, which could either be a symptom of the first failure case (right example), or a standalone issue (middle example).

To address these problems, we developed a method of ranking object detections with the confidence of the visual search system, as we found that detector confidence alone is not sufficient to suppress poor detections. In particular we looked into three signals returned by our visual search system: visual Hamming distance of the top result (from our 4096 dimensional binarized convnet features), top annotation score for annotations returned by our visual search system (aggregated tf-idf scored annotations from the visual search results), and category conformity (maximum portion of visual search results that are labeled with the same category). We list our results in Table ~\ref{tbl:spotlight_eng}, where we impose a minimum threshold on each signal. Our best variation results in a 4.9\% increase in user engagement on Flashlight. We found that category conformity in particular was critical in our improved engagement with Flashlight when object detection was enabled. A low category conformity score indicates irrelevant visual search results which we use as a proxy to suppress both of the error types shown in Figure \ref{fig:object_fp}.


%% file: 3.4-future.tex
\begin{figure}
	\centering \includegraphics[width=3.2 in]{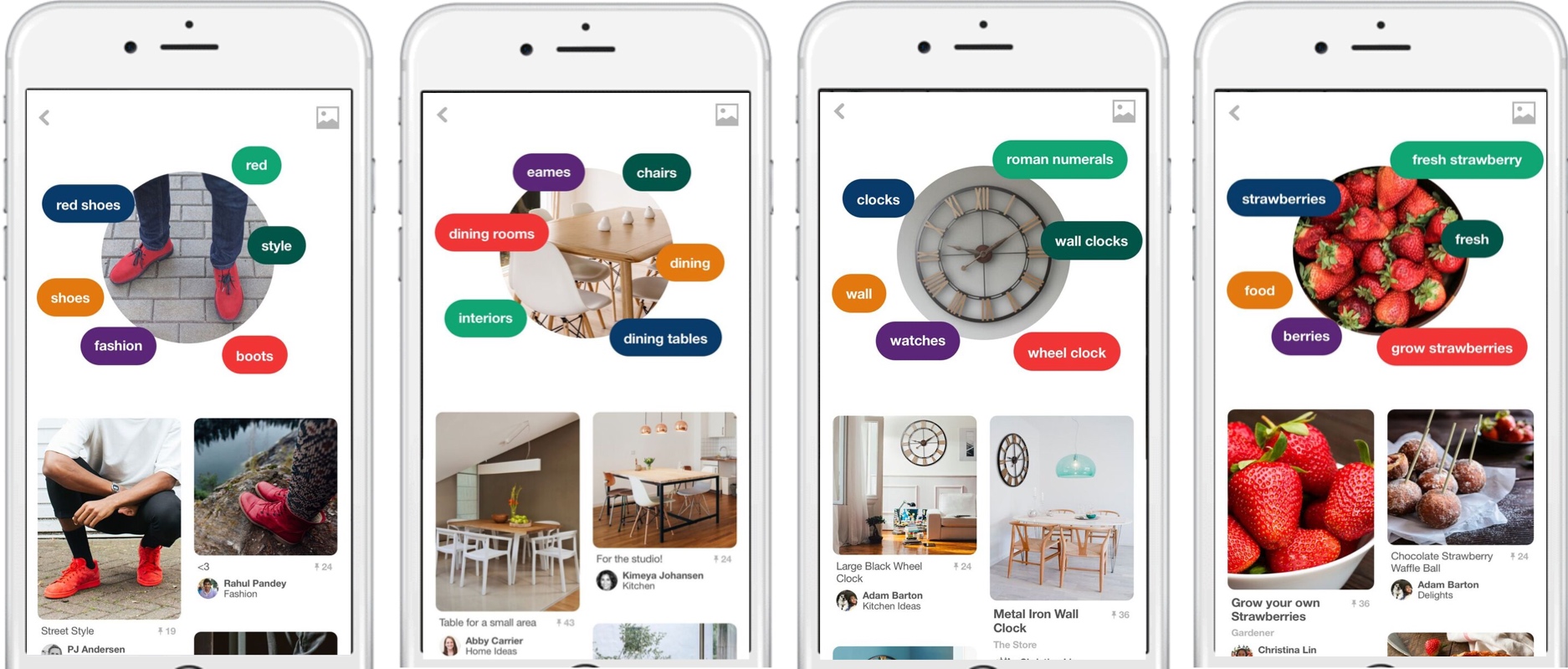}
	\caption{Pinterest Lens: user can take a picture on their phone camera to get objects and ideas related to the photo (e.g. a strawberry image can lead to chocolate strawberry recipes).}
	\label{fig:lens}
\end{figure}

\section{Pinterest Lens}
\label{sec:lens}


Pinterest Lens, as shown in Figure~\ref{fig:lens}, is a new discovery experience accessed via a mobile phone camera. Unlike Flashlight, Lens is not optimized to return visually similar results, but instead is implemented to return a diverse set of engaging results semantically relevant to the query. For example, a photo of blueberries would yield not only visually similar blueberries, the results may also include recipes for various types of blueberry desserts. 
In this section, we describe the technology used in Lens, and will report metrics in a future publication, as the product launch was very recent at the time of this writing.

\begin{figure}
	\centering \includegraphics[width=3.2 in]{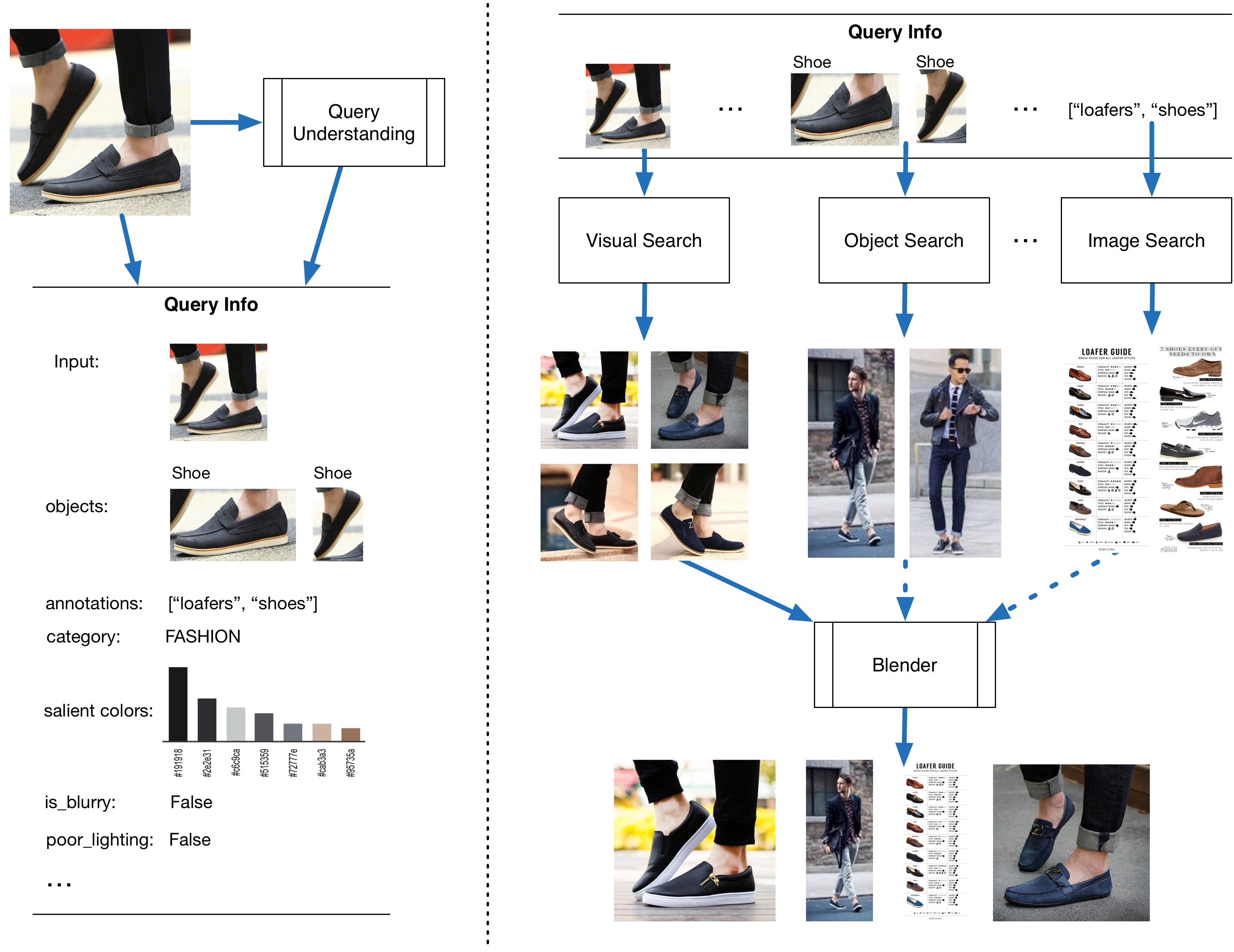}
	\caption{Lens is divided into two components. The query understanding layer (left) first computes visual features. The blending layer (right) then fetches results from multiple content sources.}
	\label{fig:lens_pipeline}
\end{figure}

The overall Lens architecture is separated into two logical components as shown in Figure~\ref{fig:lens_pipeline}. 
The first component is the query understanding layer where we returned a variety of visual and semantic features for the input image such as annotations, objects, and salient colors. 
The second component is result blending as Lens results come from multiple sources. Visually similar results come from Flashlight, semantically similar results are computed using the annotations to call Pinterest image search, and contextual results (e.g. living room designs containing the chair in the photo taken by the user's camera) come from our object search system explained below.
Not all sources are triggered per request. The blender will dynamically change blending ratios and content sources based on our derived features from the query understanding layer. For instance, image search would not be triggered if our annotations are low confidence.

\subsubsection*{Object Search}

\begin{figure}
	\centering \includegraphics[width=3 in]{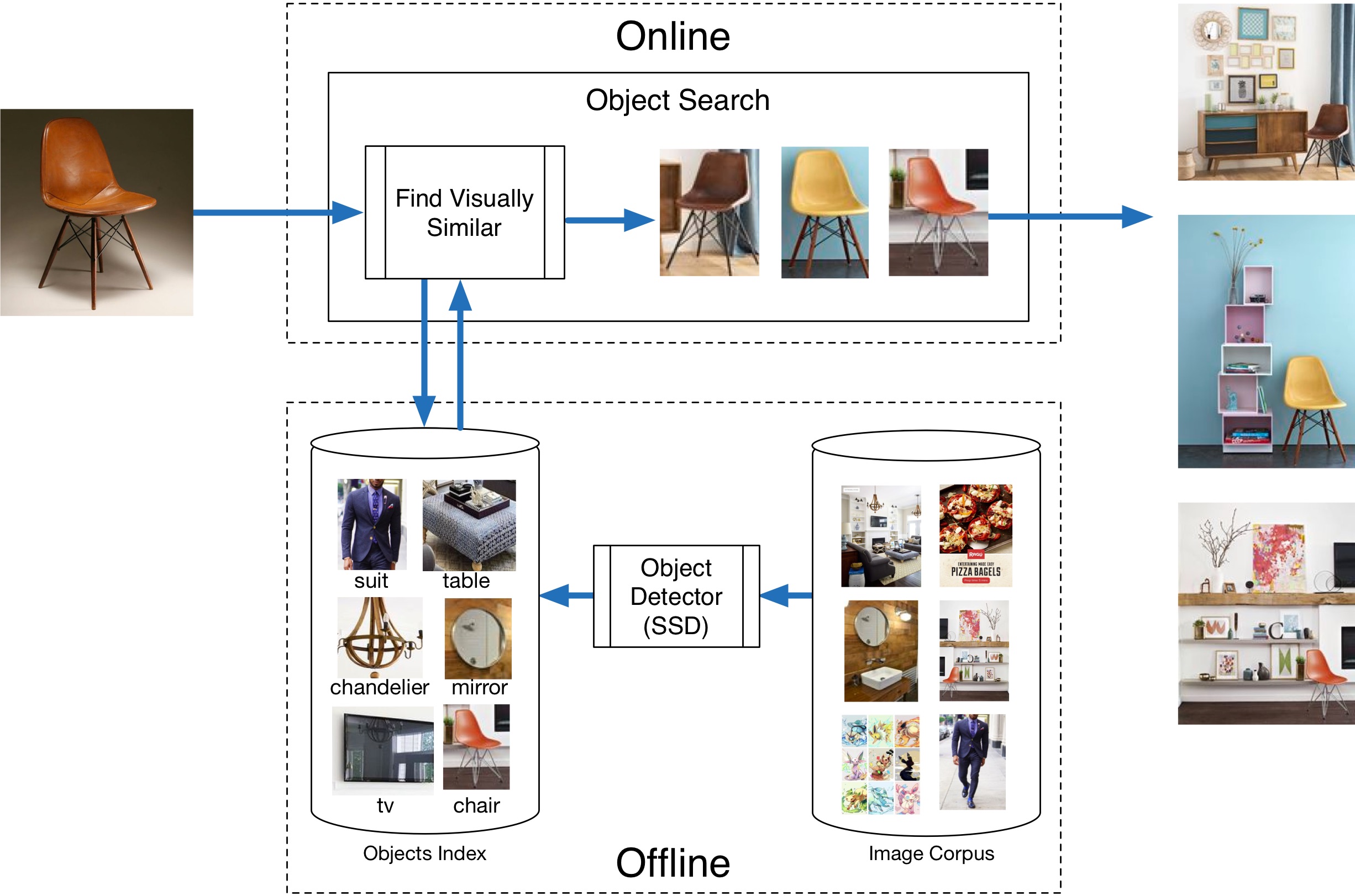}
	\caption{Given a query object, we find visually similar objects contained in larger scenes (whole images) by indexing object embeddings.}
	\label{fig:rf_pipeline}
\end{figure}

Object search is a visual search system where instead of indexing only whole images as per traditional systems, we also index objects. One use case of such a system is to retrieve results which \textit{contain} a queried object. For example, in Figure~\ref{fig:reverse}, if a user takes a picture of a Roman numeral clock, they may be interested in furniture which complements this clock or living room designs containing this clock.

To build this system, we use our SSD object detector, as described in Section~\ref{sec:detection}, to extract objects from billions of images on Pinterest for indexing. This results in a corpus with more than a billion objects. SSD is also run during query time on the given input image as object search is only triggered if the query image contains an object. Given a query object, we compute visually similar objects and then map these objects back to their whole images (scenes) to return to the user. Figure~\ref{fig:rf_pipeline} describes an end-to-end view of the pipeline. The idea of matching query with objects was previously explored in~\cite{jing15}~\cite{Bell:2015}; to the best of our knowledge this is the first production system to serve this class of recommendations.


%% file: 4-conclusion.tex
\section{Conclusions}
\label{sec:conclusion}

This paper presents an overview of our visual discovery engine powering various visual discovery experiences at Pinterest, and shares the rationales behind our technical and product decisions such as the use of binarized features, object detection, and interactive user interfaces. By sharing our experiences, we hope visual search becomes more widely incorporated into today's commercial applications.

\section{Acknowledgements}

Visual discovery is a collaborative effort at Pinterest. We'd like to thank Maesen Churchill, Jamie Favazza,  Naveen Gavini,  Yiming Jen, Eric Kim, David Liu, Vishwa Patel, Albert Pereta, Steven Ramkumar, Mike Repass,  Eric Sung, Sarah Tavel, Michelle Vu, Kelei Xu, Jiajing Xu, Zhefei Yu, Cindy Zhang, and Zhiyuan Zhang for the collaboration on the product launch, Hui Xu, Vanja Josifovski and Evan Sharp for the support. 

Additionally, we'd like to thank David Tsai, Stephen Holiday, Zhipeng Wu and Yupeng Liao for contributing to the visual search architecture at VisualGraph prior to the acquisition by Pinterest.